\documentclass[11pt,a4paper]{article}
\usepackage[hyperref]{emnlp-ijcnlp-2019}
\usepackage{times}
\usepackage{latexsym}
\usepackage{booktabs}
\usepackage{url}
\usepackage{comment}
\usepackage{multicol}
\usepackage{amsmath}
\usepackage{soul}
\usepackage{graphicx}
\usepackage{subcaption}
\aclfinalcopy 

\title{Mask-Predict: Parallel Decoding of\\
Conditional Masked Language Models}

\author{
Marjan Ghazvininejad$^{*}$ \qquad Omer Levy$^{*}$ \qquad Yinhan Liu\thanks{$^{*}$Equal contribution, sorted alphabetically.} \qquad Luke Zettlemoyer\\
Facebook AI Research\\
Seattle, WA
}

\date{}

\begin{document}
\maketitle

\begin{abstract}

Most machine translation systems generate text autoregressively from left to right.
We, instead, use a masked language modeling objective to train a model to predict any subset of the target words, conditioned on both the input text and a partially masked target translation.
This approach allows for efficient iterative decoding, where we first predict all of the target words non-autoregressively, and then repeatedly mask out and regenerate the subset of words that the model is least confident about.
By applying this strategy for a constant number of iterations, our model improves state-of-the-art performance levels for non-autoregressive and parallel decoding translation models by over 4 BLEU on average.
It is also able to reach within about 1 BLEU point of a typical left-to-right transformer model, while decoding significantly faster.\footnote{Our code is publicly available at:\\ \href{https://github.com/facebookresearch/Mask-Predict}{\color{blue}{https://github.com/facebookresearch/Mask-Predict}}}
\end{abstract}

\section{Introduction}

Most machine translation systems use sequential decoding strategies where words are predicted one-by-one.
In this paper, we present a model and a parallel decoding algorithm which, for a relatively small sacrifice in performance, can be used to generate translations in a \emph{constant} number of decoding iterations.

We introduce conditional masked language models (CMLMs), which are encoder-decoder architectures trained with a masked language model objective \cite{devlin2018,lample2019}. 
This change allows the model to learn to predict, in parallel, any arbitrary subset of masked words in the target translation. 
We use transformer CMLMs, where the decoder's self attention \cite{vaswani2017} can attend to the entire sequence (left and right context) to predict each masked word.
We train with a simple masking scheme where the number of masked target tokens is distributed uniformly, presenting the model with both easy (single mask) and difficult (completely masked) examples.
Unlike recently proposed insertion models \cite{gu2019,stern2019}, which treat each token as a separate training instance, CMLMs can train from the entire sequence in parallel, resulting in much faster training.

We also introduce a new decoding algorithm, \emph{mask-predict}, which uses the order-agnostic nature of CMLMs to support highly parallel decoding.
Mask-predict repeatedly masks out and re-predicts the subset of words in the current translation that the model is least confident about, in contrast to recent parallel decoding translation approaches that repeatedly predict the entire sequence~\cite{lee2018}.
Decoding starts with a completely masked target text, to predict all of the words in parallel, and ends after a constant number of mask-predict cycles.
This overall strategy allows the model to repeatedly reconsider word choices within a rich bi-directional context and, as we will show, produce high-quality translations in just a few cycles.

Experiments on benchmark machine translation datasets show the strengths of mask-predict decoding for transformer CMLMs. With just 4 iterations, BLEU scores already surpass the performance of the best non-autoregressive and parallel decoding models.\footnote{We use the term ``parallel decoding'' to refer to the family of approaches that can generate the entire target sequence in parallel. These are often referred to as ``non-autoregressive'' approaches, but both iterative refinement \cite{lee2018} and our mask-predict approach condition on the model's past predictions.}

With 10 iterations, the approach outperforms the current state-of-the-art parallel decoding model \cite{lee2018} by gaps of 4-5 BLEU points on the WMT'14 English-German translation benchmark, and up to 3 BLEU points on WMT'16 English-Romanian, but with the same model complexity and decoding speed. 
When compared to standard autoregressive transformer models, CMLMs with mask-predict offer a trade-off between speed and performance, trading up to 2 BLEU points in translation quality for a 3x speed-up during decoding.

\section{Conditional Masked Language Models}
\label{sec:cmlm}

A conditional masked language model (CMLM) predicts a set of target tokens $Y_{mask}$ given a source text $X$ and part of the target text $Y_{obs}$.
It makes the strong assumption that the tokens $Y_{mask}$ are conditionally independent of each other (given $X$ and $Y_{obs}$), and predicts the individual probabilities $P(y | X, Y_{obs})$ for each $y \in Y_{mask}$. 
Since the number of tokens in $Y_{mask}$ is given in advance, the model is also implicitly conditioning on the length of the target sequence $N = |Y_{mask}| + |Y_{obs}|$.

\subsection{Architecture}

We adopt the standard encoder-decoder transformer for machine translation \cite{vaswani2017}: a source-language encoder that does self-attention, and a target-language decoder that has one set of attention heads over the encoder's output and another set for the target language (self-attention).
In terms of parameters, our architecture is identical to the standard one.
We deviate from the standard decoder by removing the self-attention mask that prevents left-to-right decoders from attending on future tokens.
In other words, our decoder is bi-directional, in the sense that it can use both left and right contexts to predict each token.

\subsection{Training Objective}

During training, we randomly select $Y_{mask}$ among the target tokens.
We first sample the number of masked tokens from a uniform distribution between one and the sequence's length, and then randomly choose that number of tokens.
Following \citet{devlin2018}, we replace the inputs of the tokens $Y_{mask}$ with a special \texttt{MASK} token.

We optimize the CMLM for cross-entropy loss over every token in $Y_{mask}$.
This can be done in parallel, since the model assumes that the tokens in $Y_{mask}$ are conditionally independent of each other.
While the architecture can technically make predictions over all target-language tokens (including $Y_{obs}$), we only compute the loss for the tokens in $Y_{mask}$.

\subsection{Predicting Target Sequence Length}
\label{sec:length_prediction}

In traditional left-to-right machine translation, where the target sequence is predicted token by token, it is natural to determine the length of the sequence dynamically by simply predicting a special \texttt{EOS} (end of sentence) token.
However, for CMLMs to predict the entire sequence in parallel, they must know its length in advance.
This problem was recognized by prior work in non-autoregressive translation, where the length is predicted with a fertility model \cite{gu2017} or by pooling the encoder's outputs into a length classifier \cite{lee2018}.

We follow \citet{devlin2018} and add a special \texttt{LENGTH} token to the encoder, akin to the \texttt{CLS} token in BERT.
The model is trained to predict the length of the target sequence $N$ as the \texttt{LENGTH} token's output, similar to predicting another token from a different vocabulary, and its loss is added to the cross-entropy loss from the target sequence.

\section{Decoding with Mask-Predict}\label{sec:decoding}

We introduce the mask-predict algorithm, which decodes an entire sequence in parallel within a constant number of cycles.
At each iteration, the algorithm selects a subset of tokens to mask, and then predicts them (in parallel) using an underlying CMLM.
Masking the tokens where the model has doubts while conditioning on previous high-confidence predictions lets the model re-predict the more challenging cases, but with more information.
At the same time, the ability to make large parallel changes at each step allows mask-predict to converge on a high quality output sequence in a sub-linear number of decoding iterations.

\begin{figure*}[th]
    \centering
    \begin{tabular}{ll}
        \toprule
        $src$ & Der Abzug der französischen Kampftruppen wurde am 20. November abgeschlossen . \\
        \midrule
        $t=0$ & The \hl{departure} \hl{of} \hl{the} \hl{French} \hl{combat} \hl{completed} \hl{completed} \hl{on} 20 November . \\
        $t=1$ & The \hl{departure} of French combat troops was \hl{completed} on \hl{20} \hl{November} . \\
        $t=2$ & The withdrawal of French combat troops was completed on November 20th . \\
        \bottomrule
    \end{tabular}
    \caption{
    An example from the WMT'14 DE-EN validation set that illustrates how mask-predict generates text. 
    At each iteration, the highlighted tokens are masked and repredicted, conditioned on the other tokens in the sequence.
    }
    \label{fig:example}
\end{figure*}

\subsection{Formal Description}

Given the target sequence's length $N$ (see Section~\ref{sec:beam_search}), 
we define two variables: the target sequence $(y_1, \ldots, y_N)$ and the probability of each token $(p_1, \ldots, p_N)$.
The algorithm runs for a predetermined number of iterations $T$, which is either a constant or a simple function of $N$.
At each iteration, we perform a \emph{mask} operation, followed by \emph{predict}.

\paragraph{Mask}
For the first iteration ($t=0$), we mask all the tokens. For later iterations, we mask the $n$ tokens with the lowest probability scores:
\begin{align*}
    Y_{mask}^{(t)} &= \arg\min_i(p_i, n)\\
    Y_{obs}^{(t)} &= Y \setminus Y_{mask}^{(t)}
\end{align*}
The number of masked tokens $n$ is a function of the iteration $t$; specifically, we use linear decay $n = N \cdot \frac{T - t}{T}$, where $T$ is the total number of iterations.
For example, if $T=10$, we will mask 90\% of the tokens at $t=1$, 80\% at $t=2$, and so forth.

\paragraph{Predict}
After masking, the CMLM predicts the masked tokens $Y_{mask}^{(t)}$, conditioned on the source text $X$ and the unmasked target tokens $Y_{obs}^{(t)}$.
We select the prediction with the highest probability for each masked token $y_i \in Y_{mask}^{(t)}$ and update its probability score accordingly:
\begin{align*}
    y_i^{(t)} &= \arg\max_w P(y_i=w | X, Y_{obs}^{(t)}) \\
    p_i^{(t)} &= \max_w P(y_i=w | X, Y_{obs}^{(t)})
\end{align*}
The values and the probabilities of unmasked tokens $Y_{obs}^{(t)}$ remain unchanged:
\begin{align*}
    y_i^{(t)} &= y_i^{(t-1)} \\
    p_i^{(t)} &= p_i^{(t-1)}
\end{align*}
We tried updating or decaying these probabilities in preliminary experiments, but found that this heuristic works well despite the fact that some probabilities are stale.

\subsection{Example}
\label{sec:decoding_example}

Figure~\ref{fig:example} illustrates how mask-predict can generate a good translation in just three iterations.

In the first iteration ($t=0$), the entire target sequence is masked ($Y_{mask}^{(0)} = Y$ and $Y_{obs}^{(0)} = \emptyset$), and is thus generated by the CMLM in a purely non-autoregressive process:
\begin{align*}
P(Y_{mask}^{(0)} | X, Y_{obs}^{(0)}) = P(Y | X)    
\end{align*}
This produces an ungrammatical translation with repetitions (``completed completed''), which is typical of non-autoregressive models due to the multi-modality problem \cite{gu2017}.

In the second iteration ($t=1$), we select 8 of the 12 tokens generated in the previous step; these token were predicted with the lowest probabilities at $t=0$.
We mask them and repredict with the CMLM, while conditioning on the 4 unmasked tokens $Y_{obs}^{(1)} = \{\text{``The''}, \text{``20''}, \text{``November''}, \text{``.''}\}$.
This results in a more grammatical and accurate translation.
Our analysis shows that this second iteration removes most repetitions, perhaps because conditioning on even a little bit of the target sequence is enough to collapse the multi-modal target distribution into a single output (Section~\ref{sec:analysis_repetitions}).

In the last iteration ($t=2$), we select the 4 of the 12 tokens that had the lowest probabilities.
Two of those tokens were predicted at the first step ($t=0$), and not repredicted at the second step ($t=1$).
It is quite common for earlier predictions to be masked at later iterations because they were predicted with less information and thus tend to have lower probabilities.
Now that the model is conditioning on 8 tokens, it is able to produce an more fluent translation; ``withdrawal'' is a better fit for describing troop movement, and ``November 20th'' is a more common date format in English.

\subsection{Deciding Target Sequence Length}
\label{sec:beam_search}

When generating, we first compute the CMLM's encoder, and then use the \texttt{LENGTH} token's encoding to predict a distribution over the target sequence's length (see Section \ref{sec:length_prediction}).
Since much of the CMLM's computation can be batched, we select the top $\ell$ length candidates with the highest probabilities, and decode the same example with different lengths in parallel.
We then select the sequence with the highest average log-probability as our result:
\begin{align*}
    \frac{1}{N}\sum \log p_i^{(T)}
\end{align*}
Our analysis reveals that translating multiple candidate sequences of different lengths can improve performance (see Section~\ref{sec:length_analysis}).

\section{Experiments}
\label{sec:experiments}

We evaluate CMLMs with mask-predict decoding on standard machine translation benchmarks.
We find that our approach significantly outperforms prior parallel decoding machine translation methods and even approaches the performance of standard autoregressive models (Section \ref{sec:quality_experiments}), while decoding significantly faster (Section \ref{sec:speed_experiments}).

\subsection{Experimental Setup}

\paragraph{Translation Benchmarks}
We evaluate on three standard datasets, WMT'14 EN-DE (4.5M sentence pairs),  WMT'16 EN-RO (610k pairs) and WMT'17 EN-ZH (20M pairs) in both directions.
The datasets are tokenized into subword units using BPE \cite{sennrich2016}.
We use the same preprocessed data as \citet{vaswani2017} and \citet{wu2019pay} for WMT'14 EN-DE and WMT'17 EN-ZH respectively, and use the data from \citet{lee2018} for WMT'16 EN-RO.
We evaluate performance with BLEU \cite{papineni2002} for all language pairs, except from EN to ZH, where we use SacreBLEU \cite{post2018}.\footnote{SacreBLEU hash: BLEU+case.mixed+lang.en-zh +numrefs.1+smooth.exp+test.wmt17+tok.zh+version.1.3.7}

\paragraph{Hyperparameters}
We follow most of the standard hyperparameters for transformers in the base configuration \cite{vaswani2017}: 6 layers per stack, 8 attention heads per layer, 512 model dimensions, 2048 hidden dimensions.
We also experiment with 512 hidden dimensions, for comparison with previous parallel decoding models \cite{gu2017,lee2018}.
We follow the weight initialization scheme from BERT \cite{devlin2018}, which samples weights from $\mathcal{N}(0, 0.02)$, initializes biases to zero, and sets layer normalization parameters to $\beta=0,\gamma=1$.
For regularization, we use $0.3$ dropout, $0.01$ $L_2$ weight decay, and smoothed cross validation loss with $\varepsilon=0.1$.
We train batches of 128k tokens using Adam \cite{kingma2015} with $\beta=(0.9, 0.999)$ and $\varepsilon=10^{-6}$.
The learning rate warms up to a peak of $5\cdot10^{-4}$ within 10,000 steps, and then decays with the inverse square-root schedule.
We trained all models for 300k steps, measured the validation loss at the end of each epoch, and averaged the 5 best checkpoints to create the final model.
During decoding, we use a beam size of $b=5$ for autoregressive decoding, and similarly use $\ell=5$ length candidates for mask-predict decoding.
We trained with mixed precision floating point arithmetic on two DGX-1 machines, each with eight 16GB Nvidia V100 GPUs interconnected by Infiniband \cite{micikevicius2018mixed}.

\paragraph{Model Distillation}
Following previous work on non-autoregressive and insertion-based machine translation \cite{gu2017,lee2018,stern2019}, we train CMLMs on translations produced by a standard left-to-right transformer model (large for EN-DE and EN-ZH, base for EN-RO).
For a fair comparison, we also train standard left-to-right base transformers on translations produced by large transformer models for EN-DE and EN-ZH, in addition to the standard baselines.
We analyze the impact of distillation in Section~\ref{sec:distillation_analysis}.

\begin{table*}[th!]
\centering
\small
\begin{tabular}{lcrllll}
\toprule
\textbf{Model} & \textbf{Dimensions} & \textbf{Iterations} & \multicolumn{2}{c}{\textbf{WMT'14}} & \multicolumn{2}{c}{\textbf{WMT'16}} \\
& \textbf{(Model/Hidden)} & & \textbf{EN-DE} & \textbf{DE-EN} & \textbf{EN-RO} & \textbf{RO-EN} \\
\midrule
NAT w/ Fertility \cite{gu2017} & 512/512 & 1~~~~~~ & 19.17 & 23.20 & 29.79 & 31.44 \\
CTC Loss \cite{libovicky2018} & 512/4096 & 1~~~~~~ & 17.68 & 19.80 & 19.93 & 24.71 \\
Iterative Refinement \cite{lee2018} 
& 512/512 & 1~~~~~~ & 13.91 & 16.77 & 24.45 & 25.73 \\
& 512/512 & 10~~~~~~ & 21.61 & 25.48 & 29.32 & 30.19 \\
\qquad (Dynamic \#Iterations) & 512/512 & ?~~~~~~ & 21.54 & 25.43 & 29.66 & 30.30 \\
\midrule
\textit{Small CMLM with Mask-Predict}
& 512/512 & 1~~~~~~ & 15.06 & 19.26 & 20.12 & 20.36 \\
& 512/512 & 4~~~~~~ & \textbf{24.17} & \textbf{28.55} & \textbf{30.00} & 30.43 \\
& 512/512 & 10~~~~~~ & \textbf{25.51} & \textbf{29.47} & \textbf{31.65} & \textbf{32.27} \\
\textit{Base CMLM with Mask-Predict}
& 512/2048 & 1~~~~~~ & 18.05 & 21.83 & 27.32 & 28.20 \\
& 512/2048 & 4~~~~~~ & \textbf{25.94} & \textbf{29.90} & \textbf{32.53} & \textbf{33.23} \\
& 512/2048 & 10~~~~~~ & \textbf{27.03} & \textbf{30.53} & \textbf{33.08} & \textbf{33.31} \\
\midrule
Base Transformer \cite{vaswani2017} & 512/2048 & $N$~~~~~~ & 27.30 & ---~--- & ---~--- & ---~--- \\
Base Transformer (Our Implementation) & 512/2048 & $N$~~~~~~ & 27.74 & 31.09 & 34.28 &  33.99 \\
Base Transformer (+Distillation) & 512/2048 & $N$~~~~~~ & 27.86 & 31.07 & ---~--- &  ---~--- \\
Large Transformer \cite{vaswani2017} & 1024/4096 & $N$~~~~~~ & 28.40 & ---~--- & ---~--- & ---~--- \\
Large Transformer (Our Implementation) & 1024/4096 & $N$~~~~~~ & 28.60 & 31.71 & ---~--- & ---~--- \\
\bottomrule
\end{tabular}
\caption{The performance (BLEU) of CMLMs with mask-predict, compared to other parallel decoding machine translation methods. The standard (sequential) transformer is shown for reference. Bold numbers indicate state-of-the-art performance among parallel decoding methods.}
\label{tab:results}
\end{table*}

\begin{table*}[th!]
\centering
\small
\begin{tabular}{lcrll}
\toprule
\textbf{Model} & \textbf{Dimensions} & \textbf{Iterations} & \multicolumn{2}{c}{\textbf{WMT'17}} \\
& \textbf{(Model/Hidden)} & & \textbf{EN-ZH} & \textbf{ZH-EN} \\
\midrule
\textit{Base CMLM with Mask-Predict}
& 512/2048 & 1~~~~~~ & 24.23& 13.64\\
& 512/2048 & 4~~~~~~ & 32.63 & 21.90\\
& 512/2048 & 10~~~~~~ & 33.19 & 23.21\\
\midrule
Base Transformer (Our Implementation) & 512/2048 & $N$~~~~~~ & 34.31 & 23.74 \\
Base Transformer (+Distillation) & 512/2048 & $N$~~~~~~ & 34.44 & 23.99 \\
Large Transformer (Our Implementation) & 1024/4096 & $N$~~~~~~ & 35.01 & 24.65 \\
\bottomrule
\end{tabular}
\caption{The performance (BLEU) of CMLMs with mask-predict, compared to the standard (sequential) transformer on WMT' 17 EN-ZH.}
\label{tab:zh_results}
\end{table*}

\subsection{Translation Quality}
\label{sec:quality_experiments}

We compare our approach to three other parallel decoding translation methods:
the fertility-based sequence-to-sequence model of \citet{gu2017},
the CTC-loss transformer of \citet{libovicky2018},
and the iterative refinement approach of \citet{lee2018}.
The first two methods are purely non-autoregressive, while the iterative refinement approach is only non-autoregressive in the first decoding iteration, similar to our approach.
In terms of speed, each mask-predict iteration is virtually equivalent to a refinement iteration.

Table~\ref{tab:results} shows that among the parallel decoding methods, our approach yields the highest BLEU scores by a considerable margin.
When controlling for the number of parameters (i.e. considering only the smaller CMLM configuration), CMLMs score roughly 4 BLEU points higher than the previous state of the art on WMT'14 EN-DE, in both directions.
Another striking result is that a CMLM with only 4 mask-predict iterations yields higher scores than 10 iterations of the iterative refinement model; in fact, only 3 mask-predict iterations are necessary for achieving a new state of the art on both directions of WMT'14 EN-DE (not shown).

The translations produced by CMLMs with mask-predict also score competitively when compared to strong transformer-based autoregressive models.
In all 4 benchmarks, our base CMLM reaches within 0.5-1.2 BLEU points from a well-tuned base transformer, a relative decrease of less than 4\% in translation quality.
In many scenarios, this is an acceptable price to pay for a significant speedup from parallel decoding.

Table~\ref{tab:zh_results} shows that these trends also hold for English-Chinese translation, in both directions, despite major linguistic differences between the two languages.

\begin{figure*}[th!]
\centering
\includegraphics[width=\textwidth]{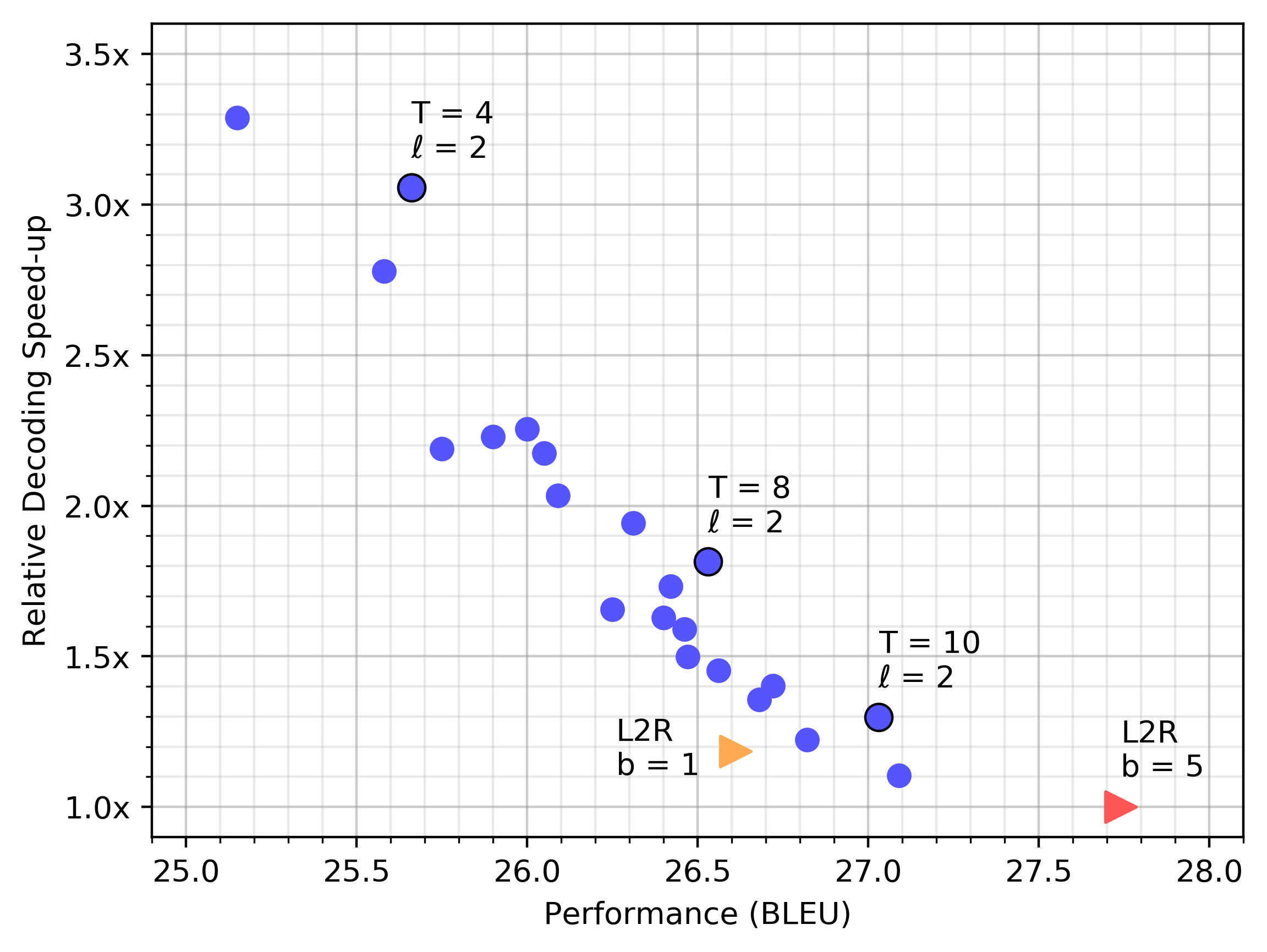}
\caption{The trade-off between speed-up and translation quality of a base CMLM with mask-predict, compared to the standard sequentially-decoded base transformer on the WMT'14 EN-DE test set, with beam sizes $b=1$ (orange triangle) and $b=5$ (red triangle). Each blue circle represents a mask-predict decoding run with a different number of iterations ($T=4,\ldots,10$) and length candidates ($\ell=1,2,3$).}
\label{fig:speed}
\end{figure*}

\subsection{Decoding Speed}
\label{sec:speed_experiments}

Because CMLMs can predict the entire sequence in parallel, mask-predict can translate an entire sequence in a constant number of decoding iterations.
Does this appealing theoretical property translate into a wall-time speed-up in practice?
By comparing the actual decoding times, we show that, for some sacrifice in performance, our parallel method can translate much faster than standard sequential transformers.

\paragraph{Setup}
As the baseline system, we use the base transformer with beam search ($b=5$) to translate WMT'14 EN-DE; we also use greedy search ($b=1$) as a faster but less accurate baseline.
For CMLMs, we vary the number of mask-predict iterations ($T=4,\ldots,10$) and length candidates ($\ell=1,2,3$).
For both models, we decode batches of 10 sentences.\footnote{The batch size was chosen arbitrarily; mask-predict can scale up to much larger batch sizes.}
For each decoding run, we measure the performance (BLEU) and wall time (seconds) from when the model and data have been loaded until the last example has been translated, and calculate the relative decoding speed-up (CMLM time / baseline time) to assess the speed-performance trade-off.

The implementation of both the baseline transformer and our CMLM is based on \texttt{fairseq} \cite{gehring2017},
which efficiently decodes left-to-right transformers by caching the state.
Caching reduces the baseline's decoding speed from 210 seconds to 128.5;
CMLMs do not use cached decoding.
All experiments used exactly the same machine and the same single GPU.

\paragraph{Results}
Figure~\ref{fig:speed} shows the speed-performance trade-off.
We see that mask-predict is versatile; on one hand, we can translate over 3 times faster than the baseline at a cost of 2 BLEU points ($T=4$, $\ell=2$), or alternatively retain a high quality of 27.03 BLEU while gaining a 30\% speed-up ($T=4$, $\ell=2$).
Surprisingly, this latter configuration outperforms an autoregressive transformer with greedy decoding ($b=1$) in both quality and speed.
We also observe that more balanced configurations (e.g. $T=8$, $\ell=2$) yield similar performance to the single-beam autoregressive transformer, but decode much faster.


\section{Analysis}
\label{sec:analysis}

To complement the quantitative results in Section~\ref{sec:experiments}, we present qualitative analysis that provides some intuition as to why our approach works and where future work could potentially improve it.

\subsection{Why Are Multiple Iterations Necessary?}
\label{sec:analysis_repetitions}

Various non-autoregressive translation models, including our own CMLM, make the strong assumption that the individual token predictions are conditionally independent of each other.
Such a model might consider two or more possible translations, A and B, but because there is no coordination mechanism between the token predictions, it could predict one token from A and another token from B.
This problem, known as the multi-modality problem \cite{gu2017}, often manifest as \emph{token repetitions} in the output when the model has multiple hypotheses that predict the same word $w$ with high confidence, but at different positions.

We hypothesize that multiple mask-predict iterations alleviate the multi-modality problem by allowing the model to condition on parts of the input, thus collapsing the multi-modal distribution into a sharper uni-modal distribution.
To test our hypothesis, we measure the percentage of repetitive tokens produced by each iteration of mask-predict as a proxy metric for the multi-modality problem.

Table~\ref{tab:repetitions} shows that, indeed, the proportion of repetitive tokens drops drastically during the first 2-3 iterations. 
This finding suggests that the first few iterations are critical for converging into a uni-modal distribution.
The decrease in repetitions also correlates with the steep rise in translation quality (BLEU), supporting the conjecture of \citet{gu2017} that multi-modality is a major roadblock for purely non-autoregressive machine translation.

\begin{table}[t]
\centering
\small
\begin{tabular}{ccccc}
\toprule
\textbf{Iterations} &  \multicolumn{2}{c}{\textbf{WMT'14 EN-DE}} &  \multicolumn{2}{c}{\textbf{WMT'16 EN-RO}}\\
 & \textbf{BLEU} & \textbf{Reps}& \textbf{BLEU} & \textbf{Reps}   \\
\midrule
$T=1$ & 18.05 &  16.72\% &  27.32 & 9.34\%  \\
$T=2$ & 22.91 &  5.40\%& 31.08 & 2.82\% \\
$T=3$ & 24.99 &  2.03\% & 32.19 & 1.26\%  \\
$T=4$ & 25.94 &  1.07\% &  32.53 & 0.87\%  \\
$T=5$ & 26.30  &  0.72\% &  32.62 & 0.61\%  \\
\bottomrule
\end{tabular}
\caption{The performance (BLEU) and percentage of repeating tokens when decoding with a different number of mask-predict iterations ($T$).}
\label{tab:repetitions}
\end{table}

\subsection{Do Longer Sequences Need More Iterations?}

A potential concern with using a constant amount of decoding iterations is that it may be effective for short sequences (where the number of iterations $T$ is closer to the output's length $N$), but insufficient for longer sequences.
To determine whether this is the case, we use \texttt{compare-mt} \cite{neubig2019} to bucket the evaluation data by target sentence length and compute the performance with different values of $T$.

Table~\ref{tab:length_vs_bleu} shows that increasing the number of decoding iterations ($T$) appears to mainly improve the performance on longer sequences.
Having said that, the performance differences across length buckets are not very large, and it seems that even 4 mask-predict iterations are enough to produce decent translations for long sequences ($40 \leq N$).

\begin{table}[t]
\centering
\small
\begin{tabular}{lccc}
\toprule
& $T = 4$ & $T = 10$ & $T = N$ \\
\midrule
$~~1 \leq N < 10$   & 21.8 & 22.4 & 22.4 \\ 
$10 \leq N < 20$	& 24.6 & 25.9 & 26.0 \\ 
$20 \leq N < 30$	& 24.9 & 26.7 & 27.1 \\ 
$30 \leq N < 40$	& 24.9 & 26.7 & 27.6 \\ 
$40 \leq N $	    & 25.0 & 27.5 & 28.1 \\ 
\bottomrule
\end{tabular}
\caption{The performance (BLEU) of base CMLM with different amounts of mask-predict iterations ($T$) on WMT'14 EN-DE, bucketed by target sequence length ($N$). Decoding with $\ell=1$ length candidates.}
\label{tab:length_vs_bleu}
\end{table}

\subsection{Do More Length Candidates Help?}
\label{sec:length_analysis}
Traditional autoregressive models can dynamically decide the length of the target sequence by generating a special \texttt{END} token when they are done, but that is not true for models that decode multiple tokens in parallel, such as CMLMs.
To address this problem, our model predicts the length of the target sequence (Section~\ref{sec:length_prediction}) and decodes multiple length candidates in parallel (Section~\ref{sec:beam_search}).
We compare our model's performance with a varying number of length candidates to its performance when conditioned on the reference (gold) target length in order to determine how accurate it is at predicting the correct length and assess the relative contribution of decoding with multiple length candidates.

Table~\ref{tab:beam_vs_bleu} shows that having multiple candidates can increase performance almost as much as conditioning on the gold length.
Surprisingly, adding too many candidates can even degrade performance.
We suspect that because CMLMs are implicitly conditioned on the target length, producing a translation that is too short (i.e. high precision, low recall) will have a high average log probability.
In preliminary experiments, we tried to address this issue by weighting the different candidates according to the model's length prediction, but this approach gave too much weight to the top candidate and resulted in lower performance.

\begin{table}
\centering
\small
\begin{tabular}{ccccc}
\toprule
\textbf{Length} & \multicolumn{2}{c}{\textbf{WMT'14 EN-DE}} & \multicolumn{2}{c}{\textbf{WMT'16 EN-RO}} \\
\textbf{Candidates} & \textbf{BLEU} & \textbf{LP} & \textbf{BLEU} & \textbf{LP} \\
\midrule
$\ell=1$   & 26.56 & 16.1\% & 32.75 & 13.8\% \\
$\ell=2$	& 27.03 & 30.6\% & 33.06 & 26.1\% \\
$\ell=3$	& \textbf{27.09} & 43.1\% & \textbf{33.11} & 39.6\% \\
$\ell=4$	& \textbf{27.09} & 53.1\% & 32.13 & 49.2\% \\
$\ell=5$	& 27.03 & 62.2\% & 33.08 & 57.5\% \\
$\ell=6$	& 26.91 & 69.5\% & 32.91 & 64.3\% \\
$\ell=7$	& 26.71 & 75.5\% & 32.75 & 70.4\% \\
$\ell=8$	& 26.59 & 80.3\% & 32.50 & 74.6\% \\
$\ell=9$	& 26.42 & 83.8\% &  32.09 & 78.3\% \\
\midrule
Gold    & 27.27 & --- & 33.20 & --- \\
\bottomrule
\end{tabular}
\caption{The performance (BLEU) of base CMLM with 10 mask-predict iterations ($T=10$), varied by the number of length candidates ($\ell$), compared to decoding with the reference target length (Gold). Length precision (LP) is the percentage of examples that contain the correct length as one of their candidates.}
\label{tab:beam_vs_bleu}
\end{table}

\subsection{Is Model Distillation Necessary?}
\label{sec:distillation_analysis}

Previous work on non-autoregressive and insertion-based machine translation reported that it was necessary to train their models on text generated by an autoregressive teacher model, a process known as distillation.
To determine CMLM's dependence on this process, we train a models on both raw and distilled data, and compare their performance.

Table~\ref{tab:distillation} shows that in every case, training with model distillation substantially outperforms training on raw data.
The gaps are especially large when decoding with a single iteration (purely non-autoregressive).
Overall, it appears as though CMLMs are heavily dependent on model distillation.

On the English-Romanian benchmark, the differences are much smaller, and after 10 iterations the raw-data model can perform comparably with the distilled model.
A possible explanation is that our teacher model was weaker for this dataset due to insufficient hyperparameter tuning.
Alternatively, it could also be the case that the English-German dataset is much noisier than the English-Romanian one, and that the teacher model essentially cleans the training data.
Unfortunately, we do not have enough evidence to support or refute either hypothesis at this time.

\begin{table}
\centering
\small
\begin{tabular}{lcccc}
\toprule
\textbf{Iterations} & \multicolumn{2}{c}{\textbf{WMT'14 EN-DE}} & \multicolumn{2}{c}{\textbf{WMT'16 EN-RO}} \\
& \textbf{Raw} & \textbf{Dist} & \textbf{Raw} & \textbf{Dist} \\
\midrule
$T=1$ & 10.64 & \textbf{18.05} & 21.22 & \textbf{27.32} \\
$T=4$ & 22.25 & \textbf{25.94} & 31.40 & \textbf{32.53} \\
$T=10$ & 24.61 & \textbf{27.03} & 32.86 & \textbf{33.08} \\
\bottomrule
\end{tabular}
\caption{The performance (BLEU) of base CMLM, trained with either raw data (Raw) or knowledge distillation from an autoregressive model (Dist).}
\label{tab:distillation}
\end{table}






\section{Related Work}


\paragraph{Training Masked Language Models with Translation Data}

Recent work by \citet{lample2019} shows that training a masked language model on sentence-pair translation data, as a pre-training step, can improve performance on cross-lingual tasks, including autoregressive machine translation.
Our training scheme builds on their work, with the following differences: we use separate model parameters for source and target texts (encoder and decoder), and we also use a different masking scheme. Specifically, we mask a varying percentage of tokens, only from the target, and do not replace input tokens with noise.
Most importantly, the goal of our work is different; we do not use CMLMs for pre-training, but to directly generate text with mask-predict decoding.

Concurrently with our work, \citet{song2019mass} extend the approach of \citet{lample2019} by using separate encoder and decoder parameters (as in our model) and pre-training them jointly in an autoregressive version of masked language modeling, although with monolingual data.
While this work demonstrates that pre-training CMLMs can improve autoregressive machine translation, it does not try to leverage the parallel and bi-directional nature of CMLMs to generate text in a non-left-to-right manner.

\paragraph{Generating from Masked Language Models}

One such approach for generating text from a masked language model casts BERT \cite{devlin2018}, a \emph{non}-conditional masked language model, as a Markov random field \cite{wang2019}.
By masking a sequence of length $N$ and then iteratively sampling a single token at each time from the model (either sequentially or in arbitrary order), one can produce grammatical examples.
While this sampling process has a theoretical justification, it also requires $N$ forward passes of the model;
mask-predict decoding, on the other hand, can produce text in a constant number of iterations.

\paragraph{Parallel Decoding for Machine Translation}

There have been several advances in parallel decoding machine translation by training non-autoregressive models. 
\citet{gu2017} introduce a transformer-based approach with explicit word fertility, and identify the multi-modality problem. 
\citet{libovicky2018} approach the multi-modality problem by collapsing repetitions with the Connectionist Temporal Classification training objective \cite{graves2006}. 
Perhaps most similar to our work is the iterative refinement approach of \citet{lee2018}, in which the model corrects the original non-autoregressive prediction by passing it multiple times through a denoising autoencoder.
A major difference is that \citet{lee2018} train their noisy autoencoder to deal with corrupt inputs by applying stochastic corruption heuristics on the training data, while we simply mask a random number of input tokens. We also show that our approach outperforms all of these models by wide margins.

\paragraph{Arbitrary Order Language Generation}

Finally, recent work has developed insertion-based transformers for arbitrary, but fixed, word order generation~\cite{gu2019,stern2019}.
While they do not decode in a constant number of iterations, \citet{stern2019} show strong results in logarithmic time.
Both models treat each token insertion as a separate training example, which cannot be computed in parallel with every other insertion in the same sequence. This makes training significantly more expensive that standard transformers (which use causal attention masking) and our CMLMs (which can predict all of the masked tokens in parallel).

\section{Conclusion}

This work introduces conditional masked language models and a novel mask-predict decoding algorithm that leverages their parallelism to generate text in a constant number of decoding iterations.
We show that, in the context of machine translation, our approach substantially outperforms previous parallel decoding methods, and can approach the performance of sequential autoregressive models while decoding much faster.
While there are still open problems, such as the need to condition on the target's length and the dependence on knowledge distillation, our results provide a significant step forward in non-autoregressive and parallel decoding approaches to machine translation.
In a broader sense, this paper shows that masked language models are useful not only for representing text, but also for \emph{generating} text efficiently.

\section*{Acknowledgements}
We thank Abdelrahman Mohamed for sharing his expertise on non-autoregressive models,
and our colleagues at FAIR for valuable feedback.

\bibliography{references}

\begin{thebibliography}{19}
\expandafter\ifx\csname natexlab\endcsname\relax\def\natexlab#1{#1}\fi

\bibitem[{Devlin et~al.(2018)Devlin, Chang, Lee, and Toutanova}]{devlin2018}
Jacob Devlin, Ming-Wei Chang, Kenton Lee, and Kristina Toutanova. 2018.
\newblock Bert: Pre-training of deep bidirectional transformers for language
  understanding.
\newblock \emph{arXiv preprint arXiv:1810.04805}.

\bibitem[{Gehring et~al.(2017)Gehring, Auli, Grangier, Yarats, and
  Dauphin}]{gehring2017}
Jonas Gehring, Michael Auli, David Grangier, Denis Yarats, and Yann~N Dauphin.
  2017.
\newblock \href {http://arxiv.org/abs/1705.03122} {{Convolutional Sequence to
  Sequence Learning}}.
\newblock \emph{ArXiv e-prints}.

\bibitem[{Graves et~al.(2006)Graves, Fern{\'a}ndez, Gomez, and
  Schmidhuber}]{graves2006}
Alex Graves, Santiago Fern{\'a}ndez, Faustino Gomez, and J{\"u}rgen
  Schmidhuber. 2006.
\newblock Connectionist temporal classification: Labelling unsegmented sequence
  data with recurrent neural networks.
\newblock In \emph{Proceedings of the 23rd international conference on Machine
  learning}, pages 369--376. ACM.

\bibitem[{Gu et~al.(2018)Gu, Bradbury, Xiong, Li, and Socher}]{gu2017}
Jiatao Gu, James Bradbury, Caiming Xiong, Victor~OK Li, and Richard Socher.
  2018.
\newblock Non-autoregressive neural machine translation.
\newblock In \emph{ICLR}.

\bibitem[{Gu et~al.(2019)Gu, Liu, and Cho}]{gu2019}
Jiatao Gu, Qi~Liu, and Kyunghyun Cho. 2019.
\newblock Insertion-based decoding with automatically inferred generation
  order.
\newblock \emph{arXiv preprint arXiv:1902.01370}.

\bibitem[{Kingma and Ba(2015)}]{kingma2015}
Diederik~P Kingma and Jimmy Ba. 2015.
\newblock Adam: A method for stochastic optimization.
\newblock In \emph{International Conference for Learning Representations}.

\bibitem[{Lample and Conneau(2019)}]{lample2019}
Guillaume Lample and Alexis Conneau. 2019.
\newblock Cross-lingual language model pretraining.
\newblock \emph{arXiv preprint arXiv:1901.07291}.

\bibitem[{Lee et~al.(2018)Lee, Mansimov, and Cho}]{lee2018}
Jason Lee, Elman Mansimov, and Kyunghyun Cho. 2018.
\newblock \href {http://www.aclweb.org/anthology/D18-1149} {Deterministic
  non-autoregressive neural sequence modeling by iterative refinement}.
\newblock In \emph{Proceedings of the 2018 Conference on Empirical Methods in
  Natural Language Processing}, pages 1173--1182, Brussels, Belgium.
  Association for Computational Linguistics.

\bibitem[{Libovick{\'y} and Helcl(2018)}]{libovicky2018}
Jind{\v{r}}ich Libovick{\'y} and Jind{\v{r}}ich Helcl. 2018.
\newblock \href {http://www.aclweb.org/anthology/D18-1336} {End-to-end
  non-autoregressive neural machine translation with connectionist temporal
  classification}.
\newblock In \emph{Proceedings of the 2018 Conference on Empirical Methods in
  Natural Language Processing}, pages 3016--3021, Brussels, Belgium.
  Association for Computational Linguistics.

\bibitem[{Micikevicius et~al.(2018)Micikevicius, Narang, Alben, Diamos, Elsen,
  Garcia, Ginsburg, Houston, Kuchaiev, Venkatesh, and
  Wu}]{micikevicius2018mixed}
Paulius Micikevicius, Sharan Narang, Jonah Alben, Gregory Diamos, Erich Elsen,
  David Garcia, Boris Ginsburg, Michael Houston, Oleksii Kuchaiev, Ganesh
  Venkatesh, and Hao Wu. 2018.
\newblock Mixed precision training.
\newblock In \emph{International Conference on Learning Representations}.

\bibitem[{Neubig et~al.(2019)Neubig, Dou, Hu, Michel, Pruthi, and
  Wang}]{neubig2019}
Graham Neubig, Zi-Yi Dou, Junjie Hu, Paul Michel, Danish Pruthi, and Xinyi
  Wang. 2019.
\newblock \href {http://arxiv.org/abs/1903.07926} {compare-mt: A tool for
  holistic comparison of language generation systems}.
\newblock In \emph{Meeting of the North American Chapter of the Association for
  Computational Linguistics (NAACL) Demo Track}, Minneapolis, USA.

\bibitem[{Papineni et~al.(2002)Papineni, Roukos, Ward, and Zhu}]{papineni2002}
Kishore Papineni, Salim Roukos, Todd Ward, and Wei-Jing Zhu. 2002.
\newblock \href {https://doi.org/10.3115/1073083.1073135} {{B}leu: a method for
  automatic evaluation of machine translation}.
\newblock In \emph{Proceedings of the 40th Annual Meeting of the Association
  for Computational Linguistics}, pages 311--318, Philadelphia, Pennsylvania,
  USA. Association for Computational Linguistics.

\bibitem[{Post(2018)}]{post2018}
Matt Post. 2018.
\newblock \href {https://doi.org/10.18653/v1/W18-6319} {A call for clarity in
  reporting {BLEU} scores}.
\newblock In \emph{Proceedings of the Third Conference on Machine Translation:
  Research Papers}, pages 186--191, Belgium, Brussels. Association for
  Computational Linguistics.

\bibitem[{Sennrich et~al.(2016)Sennrich, Haddow, and Birch}]{sennrich2016}
Rico Sennrich, Barry Haddow, and Alexandra Birch. 2016.
\newblock \href {https://doi.org/10.18653/v1/P16-1162} {Neural machine
  translation of rare words with subword units}.
\newblock In \emph{Proceedings of the 54th Annual Meeting of the Association
  for Computational Linguistics (Volume 1: Long Papers)}, pages 1715--1725,
  Berlin, Germany. Association for Computational Linguistics.

\bibitem[{Song et~al.(2019)Song, Tan, Qin, Lu, and Liu}]{song2019mass}
Kaitao Song, Xu~Tan, Tao Qin, Jianfeng Lu, and Tie-Yan Liu. 2019.
\newblock Mass: Masked sequence to sequence pre-training for language
  generation.
\newblock \emph{arXiv preprint arXiv:1905.02450}.

\bibitem[{Stern et~al.(2019)Stern, Chan, Kiros, and Uszkoreit}]{stern2019}
Mitchell Stern, William Chan, Jamie Kiros, and Jakob Uszkoreit. 2019.
\newblock Insertion transformer: Flexible sequence generation via insertion
  operations.
\newblock \emph{arXiv preprint arXiv:1902.03249}.

\bibitem[{Vaswani et~al.(2017)Vaswani, Shazeer, Parmar, Uszkoreit, Jones,
  Gomez, Kaiser, and Polosukhin}]{vaswani2017}
Ashish Vaswani, Noam Shazeer, Niki Parmar, Jakob Uszkoreit, Llion Jones,
  Aidan~N Gomez, {\L}ukasz Kaiser, and Illia Polosukhin. 2017.
\newblock Attention is all you need.
\newblock In \emph{Advances in Neural Information Processing Systems}, pages
  5998--6008.

\bibitem[{Wang and Cho(2019)}]{wang2019}
Alex Wang and Kyunghyun Cho. 2019.
\newblock Bert has a mouth, and it must speak: Bert as a markov random field
  language model.
\newblock \emph{arXiv preprint arXiv:1902.04094}.

\bibitem[{Wu et~al.(2019)Wu, Fan, Baevski, Dauphin, and Auli}]{wu2019pay}
Felix Wu, Angela Fan, Alexei Baevski, Yann~N Dauphin, and Michael Auli. 2019.
\newblock Pay less attention with lightweight and dynamic convolutions.
\newblock \emph{International Conference on Learning Representations}.

\end{thebibliography}
\bibliographystyle{acl_natbib}

\end{document}